\documentclass[conference]{IEEEtran}
\usepackage{times}

\usepackage[numbers]{natbib}
\usepackage{multicol}
\usepackage[bookmarks=true]{hyperref}
\usepackage{easy-todo}
\usepackage{graphicx}
\usepackage{booktabs}
\usepackage{dirtree}
\usepackage{afterpage}
\usepackage{amsmath}
\makeatletter
\renewcommand{\DTcomment}[1]{%
  \hspace*{\fill}%
  \makebox[0.43\textwidth][l]{\DTstylecomment #1}%
}
\makeatother
\usepackage{caption}
\usepackage{booktabs}

\begin{document}

% paper title
\title{ForVis: An In-Field Dataset and Benchmark for VIO Using Under-Canopy UAV Flights in Forests}

% You will get a Paper-ID when submitting a pdf file to the conference system
\author{%
\parbox{\textwidth}{%
\centering
Arman Kiani$^{1}$, Masoud Ataei$^{1}$, Elvis Gyaase$^{1}$, Jeffrey Eiyike$^{1}$\\
Aaron Weiskittel$^{2}$, Prabuddha Chakraborty$^{1}$, Vikas Dhiman$^{1}$\\[0.6em]
{\small
$^{1}$Department of Electrical and Computer Engineering, University of Maine, Orono, ME 04469, USA\\
$^{2}$School of Forest Resources, University of Maine, Orono, ME 04469, USA\\[0.3em]
\texttt{\{arman.kiani, masoud.ataei, elvis.gyaase, jeffrey.eiyike, aaron.weiskittel, prabuddha, vikas.dhiman\}@maine.edu}
}
}}
\maketitle

\begin{abstract}
Visual-inertial Simultaneous Localization and Mapping (VI-SLAM) for UAVs remains difficult to evaluate in real forest environments, where motion, illumination changes, repetitive vegetation, and vibration can all affect estimation.
We present ForVis, an in-field dataset and benchmark for evaluating VI-SLAM during UAV flight in forest environments. 
The dataset contains twelve flights across open meadow, above-canopy, and under-canopy conditions in each environment. 
In total, it provides 563.8s of flight over 1096.8m of trajectory, recorded simultaneously with an Intel RealSense D435i and an OAK-D Pro Wide together with inertial and flight-controller data. 
We benchmark seven open-source VI-SLAM systems over 504 runs.
The results show that sensor choice has a larger effect on trajectory error than the spread between algorithms: all seven methods achieve lower median error on the OAK-D Pro than on the D435i.
%Flight sequence difficulty is reflected more strongly in failure rate (loss of tracked features) than in the average translation error of successful runs, while above-canopy flights provide the clearest separation between methods.
ForVis is intended to support evaluation of speed, accuracy and robustness for VI-SLAM in challenging forest flight.
\end{abstract}

\begin{center}
\footnotesize
\href{https://drive.google.com/drive/folders/1T3nJRqQwyy-msXUD_-oBVeV6XnQXUZ76?usp=drive_link}{\textbf{ForVis Dataset Download}\footnote{\url{https://drive.google.com/drive/folders/1T3nJRqQwyy-msXUD_-oBVeV6XnQXUZ76}}}
\end{center}

\IEEEpeerreviewmaketitle

\begin{figure*}[!t]
    \centering
    \includegraphics[width=\textwidth]{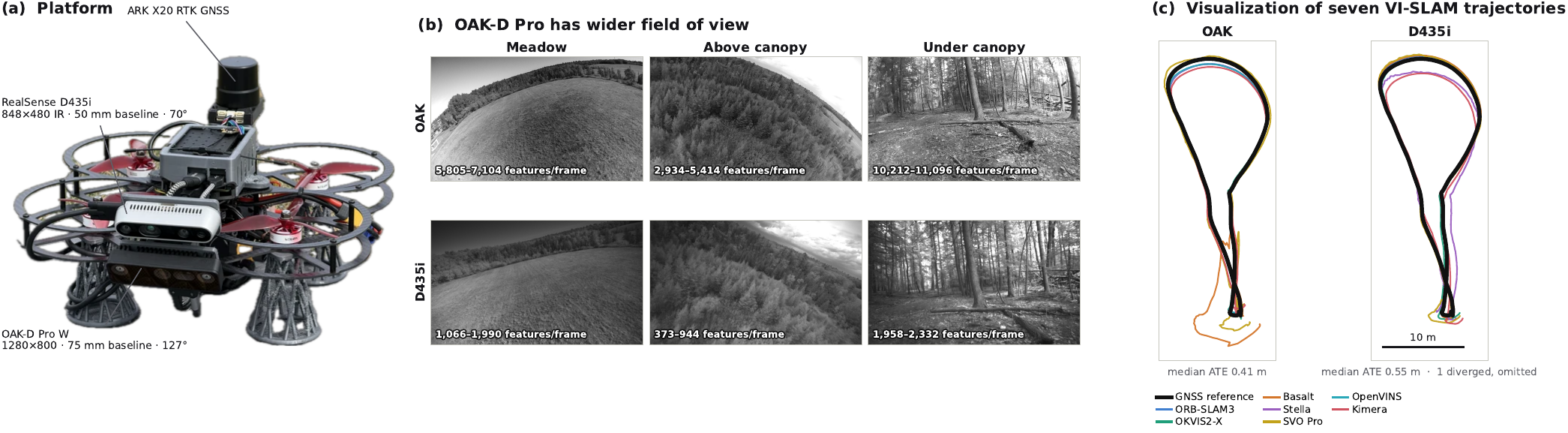}
    \caption{ForVis records visual-inertial data from two sensors, OAK-D Pro and RealSense D435i, on a UAV. (a) The platform is an ARK-RIG5 inspired drone with an OAK-D Pro W beneath a RealSense D435i on a 5-inch quadrotor, with an ARK X20 RTK GNSS receiver above for ground truth. (b) The same three scenes through both sensors in each of the three environments, annotated with the median FAST corner count per frame — the forest interior is the most feature-rich environment in the dataset and the canopy top the poorest, on both sensors. (c) Seven visual-inertial methods on \texttt{mdw\_1} sequence, aligned to the RTK reference (black). Every method has lower performance on D435i than OAK-D Pro.}
    \label{fig:teaser}
\end{figure*}

\section{Introduction}
% 1. What is the problem, why is it important?
% 2. How is it typically solved? Related work?
% 3. What is missing in the related work?
% 4. How do you plan to address it?
% 5. Did it get addressed or not?

% Start with a sentence that say everyone agrees.
% Why we need under-canopy flights?
% What does "under-canopy forest" mean? Under-canopy flight in a forest.
Under-canopy forests are complex, unstructured, and often GNSS-degraded environments where Unmanned Aerial Vehicles (UAVs) must navigate among trees, branches, dense vegetation, uneven terrain, and rapidly changing illumination.
Operating below the canopy enables applications that above-canopy sensing cannot fully support, such as detailed forest inventory, close-range inspection, search and rescue, environmental monitoring, and autonomous exploration.

Historically, many autonomous UAVs operating in complex outdoor and forest environments have relied on LiDAR for localization, mapping, and obstacle perception~\cite{prabhu2024uavsforestry}.
LiDAR provides direct geometric measurements and is relatively insensitive to illumination changes, making it particularly suitable for environments containing trees and other geometric structures.
For example, Liu et al.~\cite{liu2022large} demonstrated large-scale autonomous flight under dense forest canopy using Semantic Localization and Mapping (SLAM) in which tree trunks and ground planes extracted from LiDAR measurements were used to constrain the UAV pose and construct a semantic representation of the forest.

Despite these advantages, a LiDAR is heavier, more power hungry, and costlier than a camera or even a depth camera, making it impractical for small UAV platforms with limited payload capacity and flight endurance. 
%On the other hand, cameras are typically lightweight, inexpensive, and information-rich sensors and can be combined with inertial measurements to provide visual-inertial odometry (VIO) and visual-inertial simultaneous localization and mapping (VI-SLAM).
Therefore, there is increasing interest in enabling autonomous flight in complex outdoor environments using primarily visual and inertial sensing.
Recent work by Laina et al.~\cite{laina2025scalable}, for example, demonstrated large-scale autonomous UAV navigation in unstructured and cluttered outdoor environments using lightweight passive visual and inertial sensors without relying on LiDAR. 
These developments provide compelling evidence that visual-inertial perception could enable smaller and more accessible autonomous aerial platforms.
However, in the field, under-canopy flights involve changing illumination, repetitive visual patterns, moving foliage, motion blur, vibration, dropped frames, aggressive rotations, and vegetation disturbed by propeller downwash.
Under such conditions, the robustness of Visual-Inertial SLAM (VI-SLAM) algorithms---
such as ORB-SLAM3~\cite{ORB-SLAM3},
Basalt-VIO~\cite{basalt}, stella\_vslam~\cite{sumikura2019openvslam},
OpenVINS~\cite{geneva2020openvins}, OKVIS2-X~\cite{boche2025okvis2}, Kimera-VIO~\cite{rosinol2020kimera}
---remains untested.
%but they also expose the estimator to failure modes that are particularly severe in forest environments.

To evaluate and compare available VI-SLAM algorithms for speed, accuracy and robustness in under-canopy flights, we need an evaluation dataset and benchmark.
There exists several standard VI-SLAM datasets, such as EuRoC~\cite{burri2016euroc}, TUM-VI~\cite{schubert2018tum}, KITTI~\cite{geiger2012we}, however they are either in indoor or in urban environments that do not capture the challenges of forest environments.
%Reliable data from under-canopy environments is important because commonly used VIO datasets do not fully capture the challenges of real UAV flight in forests.
Forest specific datatsets such as FinnForest~\cite{ali2020finnforest} target VI-SLAM from a ground vehicle perspective.
Additionally, Wild-Places~\cite{knights2023wild} and the Oxford Forest dataset~\cite{oh2024evaluation} focus mainly on LiDAR-based place recognition using handheld or backpack platforms.
None of the available datasets, however, capture under-canopy UAV flight data in real forests.
%GNSS (Global Navigation Satellite System) quality can also change significantly as the UAV moves through areas with different canopy density. 
%A dataset collected during real forest flights is therefore valuable for evaluating and comparing VIO algorithms perform in the field and for understanding when and why their performance begins to degrade.

To address this need, we introduce Forest Vision (ForVis) dataset, an under-canopy UAV dataset designed for benchmarking VI-SLAM algorithms as shown in Fig.~\ref{fig:teaser}. 
The main components of the dataset are: (1) synchronized visual and inertial data from two onboard camera systems together with PX4 ULog flight-controller logs from real under-canopy UAV flights; (2) RTK fixed GNSS positioning for trajectories collected in open areas; and (3) evaluation of VI-SLAM trajectories using different reference strategies depending on the environment. In open areas, RTK fixed positioning is used as the ground-truth reference. Under the forest canopy, where reliable RTK positioning is unavailable, trajectory consistency is instead evaluated using loop-closure constraints. 

% In summary, our main contributions are: 1) a real-world UAV dataset and benchmark for evaluating VI-SLAM in meadow, above-canopy, and under-canopy forest flight, 2) a comparison of seven open-source SLAM/VIO methods on OAK-D Pro W and RealSense D435i sensors, including accuracy, completion rate, and computational cost, and 3) an analysis showing that under-canopy flight remains challenging across methods and that sensor configuration and calibration can strongly influence trajectory error and failure behavior.

%... \todoi{what do we find?} \ARM{honestly I found out that camera is more important than methods and there is no method that can be called as the best method. Does it count as a contribution?}

% Removed for space
% \input{tex/relatedwork.tex}

\begin{table*}[t]
  \centering
  \caption{The twelve benchmark sequences: three environments, with six sequences
  under canopy. 
  The roll/pitch column is the total attitude change and yaw is cumulative
  heading change, both over the whole flight. 
  Frames is the percentage of image frames saved relative the requested rate (30 FPS for D435i and 40 FPS for OAK-D), reported for whichever sensors delivered fewer. 
  Closure is the return-to-start closure: the RTK reference's own residual for the open-sky sequences, and the pilot's measured pad return under canopy, measured manually ($^\dagger$, $\pm5$\,cm), where the receiver drifts further than the aircraft moves.}
  \label{tab:datasets}
  \begin{tabular}{lrrrrrrrrrrll}
    \toprule
    Sequence & Roll/Pitch & Yaw & Duration & Path & Alt.\ range & $v_{\max}$ & $\lVert\omega\rVert$ p99 & Images & Frames & Size & Closure & Ground truth \\
             & [$^\circ$] & [$^\circ$] & [s] & [m] & [m] & [m/s] & [$^\circ$/s] & & [\%] & [GB] & [m] & \\
    \midrule
    \texttt{mdw\_1} & 31/32 & 151 & 39.7 & 74.0 & 7.0 & 4.72 & 93 & 7\,281 & 94.4 & 1.8 & 0.05 & RTK fixed, 2.0\,cm \\
    \texttt{mdw\_2} & 52/37 & 140 & 34.7 & 70.4 & 13.8 & 7.99 & 81 & 6\,159 & 87.3 & 1.5 & 0.10 & RTK fixed, 2.0\,cm \\
    \texttt{mdw\_3} & 69/70 & 427 & 39.1 & 66.6 & 5.6 & 5.49 & 158 & 7\,086 & 91.8 & 1.8 & 0.10 & RTK fixed, 2.0\,cm \\
    \midrule
    \texttt{abv\_1} & 35/41 & 386 & 54.5 & 93.8 & 10.8 & 4.69 & 117 & 9\,837 & 91.8 & 2.0 & 0.02 & RTK fixed, 2.0\,cm \\
    \texttt{abv\_2} & 54/57 & 464 & 52.6 & 108.4 & 14.1 & 8.64 & 98 & 9\,585 & 93.6 & 2.1 & 0.05 & RTK fixed, 2.0\,cm \\
    \texttt{abv\_3} & 51/60 & 365 & 50.1 & 187.3 & 30.1 & 10.09 & 99 & 9\,174 & 91.4 & 2.0 & 0.28 & RTK fixed, 8.0\,cm \\
    \midrule
    \texttt{can\_1} & 43/43 & 119 & 38.8 & 94.0 & 21.7 & 7.01 & 91 & 7\,256 & 93.2 & 2.2 & 0.23$^\dagger$ & GNSS unusable \\
    \texttt{can\_2} & 14/24 & 202 & 42.4 & 58.0 & 10.5 & 3.55 & 134 & 7\,660 & 90.8 & 2.1 & 0.32$^\dagger$ & GNSS unusable \\
    \texttt{can\_3} & 34/37 & 154 & 47.0 & 84.0 & 19.3 & 5.75 & 108 & 9\,404 & 93.3 & 2.8 & 0.23$^\dagger$ & GNSS unusable \\
    \texttt{can\_4} & 39/56 & 168 & 58.1 & 90.4 & 17.5 & 4.14 & 93 & 10\,880 & 94.8 & 3.2 & 0.20$^\dagger$ & GNSS unusable \\
    \texttt{can\_5} & 50/55 & 260 & 54.0 & 81.6 & 16.4 & 6.25 & 99 & 10\,084 & 90.8 & 3.1 & 0.12$^\dagger$ & GNSS unusable \\
    \texttt{can\_6} & 56/42 & 326 & 52.7 & 88.3 & 22.2 & 6.23 & 111 & 11\,405 & 94.2 & 3.3 & 0.35$^\dagger$ & GNSS unusable \\
    \midrule
    Total & & & 563.8 & 1096.8 & & & & 105\,811 & & 28.1 & & \\
    \bottomrule
  \end{tabular}
\end{table*}

\section{ForVis Dataset}

%\subsection{Dataset Statistics}

ForVis dataset contains twelve flight sequences spanning open meadow (\texttt{mdw\_*}), above-canopy (\texttt{abv\_*}), and under-canopy (\texttt{can\_*}) environments.
In total, the dataset contains 563.8~s of flight over 1096.8~m of aerial track, producing 105,811 stereo image pairs and 28.1~GB of data.

Table~\ref{tab:datasets} summarizes the twelve benchmark sequences. 
% Difficulty labels were assigned within each environment from the overall flight dynamics rather than subjective assessment, considering translation speed, attitude excursion, angular motion,  and cumulative heading change together. 
The roll and pitch column reports the total attitude change over a sequence, yaw reports the cumulative heading change, and ``$\lVert\omega\rVert$ p99'' reports the 99th percentile of the body angular-velocity magnitude, which reduces sensitivity to isolated spikes as compared to the maximum angular velocity.
The table also reports sequence duration, path length, altitude range, maximum speed, recorded stereo pairs, frame yield, dataset size, and return-to-start closure. 
Here return-to-start closure is the difference between estimated change between start and end position computed by the VI-SLAM algorithm and manually-measured difference between those positions for the under canopy sequences, and measured by Real-time Kinematic (RTK) GNSS for other sequences.
%Because no single motion metric increases monotonically with difficulty, the labels reflect the combined motion characteristics of each flight.
%The percentile is reported rather than the maximum because the airframe carries a propeller imbalance whose vibration reaches the gyro: single-sample peaks run to 525 °/s, several times any commanded rotation, and ranking the sequences by that maximum orders them by rotor balance rather than by flight.
%The p99 is stable across the set at 81–158 °/s against a median of 14 °/s.

RTK GNSS uses correction data from a fixed reference station to reduce positioning errors and provide centimeter-level positioning. RTK-fixed ground truth is available for the six meadow and above canopy sequences, with reported horizontal noise of approximately 2~cm, except for one sequence at 8~cm (\texttt{abv\_3}). Under canopy, reliable RTK lock was not maintained and reported GNSS noise increased to 1.64--2.78~m. As an additional reference, the measured return-to-launch closure of 12--35~cm indicates the discrepancy between the estimated and initial launch positions after the system returned to its starting point. Therefore, the return-to-launch closure was used as  a proxy for accuracy in under canopy sequences instead of GNSS.

\begin{table*}[t]
\centering
\caption{Accuracy and computational cost for the two sensors. ATE is the median over the
sequences of each environment, and path error is the ATE expressed as a fraction of the reconstructed
path length. The superscript marks ${}^1$, ${}^2$ indicate median over fewer sequences than three and
``---'' indicates that none of the sequences were tracked completely. The under-canopy columns are scored
against a meter-grade reference and are not comparable with the other four. Cost columns
are computed over the ten sequences measured under a strictly serial condition (one
container, eight physical cores, nothing else running); the two sequences added on
4 September were measured with two workers and are excluded from cost, though not from
accuracy. $^{*}$SVO Pro Pro is fed at capture rate, so its real-time factor is fixed by
construction rather than measured.}
\label{tab:cost}

\resizebox{\textwidth}{!}{%
\begin{tabular}{lccccccccccc}
\toprule
& \multicolumn{2}{c}{Meadow}
& \multicolumn{2}{c}{Above canopy}
& \multicolumn{2}{c}{Under canopy}
& & & & & \\
\cmidrule(lr){2-3}
\cmidrule(lr){4-5}
\cmidrule(lr){6-7}

Method &
\shortstack{ATE\\(m)} &
\shortstack{Path\\error} &
\shortstack{ATE\\(m)} &
\shortstack{Path\\error} &
\shortstack{ATE\\(m)} &
\shortstack{Path\\error} &
\shortstack{Avg. runtime\\(s)} &
\shortstack{RTF\\wall / data} &
\shortstack{CPU mean\\(\%)} &
\shortstack{Peak mem.\\(MB)} &
\shortstack{Seqs\\completed} \\
\midrule

\multicolumn{12}{l}{\textbf{OAK-D Pro W}} \\
\midrule
\texttt{orb\_slam3} & 0.58 & 0.77\% & 3.72$^{2}$ & 2.33\%$^{2}$ & \textbf{1.75} & \textbf{2.25}\% & 65.0 & 1.45 & 238 & 774 & 11/12 \\
\texttt{okvis2\_x} & \textbf{0.28} & \textbf{0.38}\% & 1.24 & 0.99\% & 1.82 & 2.06\% & 65.7 & 1.43 & 234 & 1016 & 12/12 \\
\texttt{basalt} & 1.71 & 1.63\% & \textbf{1.03} & \textbf{0.82}\% & 1.96 & 2.12\% & 32.0 & 0.70 & 725 & 241 & 12/12 \\
\texttt{stella\_vslam} & 1.10 & 1.52\% & 2.54 & 2.10\% & 1.88 & 2.15\% & 178.5 & 3.89 & 174 & 811 & 12/12 \\
\texttt{svo\_pro\_open} & 0.62 & 0.37\% & 1.31 & 0.69\% & 1.82 & 1.88\% & 64.6 & 1.42$^{*}$ & 94 & 354 & 12/12 \\
\texttt{openvins} & 0.38 & 0.49\% & 1.54 & 1.47\% & 1.88 & 2.17\% & 25.5 & 0.56 & 102 & 363 & 12/12 \\
\texttt{kimera\_vio} & 0.86 & 1.13\% & 8.90$^{2}$ & 11.49\%$^{2}$ & 2.32 & 3.08\% & 121.9 & 2.63 & 188 & 7250 & 11/12 \\
\midrule
\multicolumn{12}{l}{\textbf{RealSense D435i}} \\
\midrule
\texttt{orb\_slam3} & 0.33 & 0.41\% & 1.00 & 0.85\% & 1.94 & 2.17\% & 51.8 & 1.14 & 144 & 718 & 12/12 \\
\texttt{okvis2\_x} & 0.42 & 0.53\% & 1.25 & 1.01\% & 2.07 & 2.21\% & 40.0 & 0.88 & 195 & 529 & 12/12 \\
\texttt{basalt} & 0.41 & 0.51\% & 1.88 & 0.89\% & 2.26 & 2.31\% & \textbf{5.7} & \textbf{0.13} & 786 & 117 & 12/12 \\
\texttt{stella\_vslam} & 2.85 & 2.89\% & 8.66$^{2}$ & 12.97\%$^{2}$ & 2.10 & 2.29\% & 31.8 & 0.70 & 195 & 404 & 11/12 \\
\texttt{svo\_pro\_open} & 0.64 & 0.67\% & 2.93$^{2}$ & 0.73\%$^{2}$ & 2.00 & 1.83\% & 64.3 & 1.42$^{*}$ & 58 & 257 & 11/12 \\
\texttt{openvins} & 0.82$^{2}$ & 0.89\%$^{2}$ & 2.67$^{2}$ & 1.79\%$^{2}$ & 2.17 & 2.29\% & 12.9 & 0.29 & 83 & 245 & 10/12 \\
\texttt{kimera\_vio} & 2.01 & 2.30\% & 10.50$^{2}$ & 13.24\%$^{2}$ & 2.14 & 2.41\% & 111.6 & 2.37 & 155 & 2612 & 11/12 \\
\bottomrule
\end{tabular}%
}
\end{table*}

\section{VI-SLAM Benchmark and Dataset Analysis}

We evaluate seven open-source algorithms: 
ORB-SLAM3~\cite{ORB-SLAM3},
Basalt-VIO~\cite{basalt}, stella\_vslam~\cite{sumikura2019openvslam},
OpenVINS~\cite{geneva2020openvins}, SVO Pro~\cite{forster2016svo},
OKVIS2-X~\cite{boche2025okvis2}, and Kimera-VIO~\cite{rosinol2020kimera}. 
Six out of the seven algorithms are visual-inertial, while Stella-VSLAM is stereo-only.
% We Repetition matters because several of these systems are non-deterministic: two of the seven produce different results from identical input, and a single run would misrepresent them in either direction.
ORB-SLAM3~\cite{ORB-SLAM3} combines feature-based stereo-inertial tracking with bundle adjustment, sparse keyframe mapping, and loop closure. 
Basalt-VIO~\cite{basalt} instead relies on optical-flow tracking and sliding-window optimization without loop closure. stella\_vslam~\cite{sumikura2019openvslam} was used in stereo-only mode as a feature-based SLAM baseline with sparse mapping and loop-closure detection, while OpenVINS~\cite{geneva2020openvins} provides a contrasting filter-based approach built around the MSCKF~\cite{mourikis2007multi} and produced deterministic results in our experiments.
Specifically, we use the stella\_vslam implementation available
\href{https://github.com/stella-cv/stella_vslam}{here}\footnote{\url{https://github.com/stella-cv/stella_vslam}}.
SVO Pro uses a semi-direct formulation, combining direct image alignment with feature matching, and was run in VIO-only mode with loop closure disabled~\cite{forster2016svo}. 
OKVIS2-X~\cite{boche2025okvis2} uses a sliding-window factor graph that tightly couples IMU pre-integration with visual reprojection errors, although it showed greater run-to-run variation and occasional divergence. Kimera-VIO~\cite{rosinol2020kimera} also uses factor-graph optimization, with an GTSAM~\cite{dellaert2012factor} backend, and only its stereo-inertial odometry component was considered here.

We evaluate all seven algorithms on all twelve sequences repeating three times with different random seeds on both sensors, giving 504 runs ($=7\times3\times2\times12$).
The mean of the repeated runs is reported for ``CPU mean'' and ``Peak mem'' in Table~\ref{tab:results}, while median is reported for the Average Translation Error (ATE) and ``Path error'' in the table.
All experiments run on the same workstation (AMD Ryzen 9 9955HX, 32~GB
RAM). 
Each method runs in a separate Docker container, pinned to eight
physical CPU cores with SMT siblings left idle and limited to 16~GB of
memory.

\subsection{Benchmark Results}

Table~\ref{tab:cost} summarizes accuracy and computational cost across the seven methods. 
Detailed per-sequence benchmark results are provided in Appendix~\ref{app:benchmark_results}.
Our evaluations yield different results for different kinds of sequences: meadow, above canopy, and under-canopy.
In meadow sequences, SVO pro  and OKVIS2-X provide the highest accuracy of 0.37\% and 0.38\%, respectively, on OAK-D, while SVO Pro is robust to changes in sensors retaining an accuracy of 0.37\% on OAK-D and 0.67\% on D435i.
% However, SVO Pro is not able to estimate scale in the absence of IMU information.
% Stell-VSLAM cannot estimate scale without IMU information.
% Whereas Kimera-VIO is the slowest with real-time factor of 1.86 and has the largest memory footprint of 2575 MB. 
Above canopy, SVO Pro gives the lowest path error on both sensors (0.69\% and 0.73\%), while several methods show substantially larger errors or incomplete sequences, particularly Stella-VSLAM and Kimera-VIO. 
Under canopy, errors are more consistently increased across methods: path error ranges from 1.88--3.08\% on OAK-D and 1.83--2.41\% on D435i, indicating that the environment remains challenging even for methods that perform well in open scenes.
On OAK-D, under-canopy path error remains around 1.9--3.1\% for all methods, higher than in the meadow, while the D435i shows roughly 1.8--2.4\% together with more incomplete sequences.
Across all sequences, Basalt and OpenVINS achieve the lowest real-time computational factors of 0.13 and 0.29. 

\subsubsection{Challenges of Under-canopy VI-SLAM}
Across the 504 benchmark runs, 472 produced a usable trajectory estimate. 
As shown in Table~\ref{tab:failure_summary}, failures occurred with both sensors but were substantially more frequent with the D435i: 10\% of its runs lost tracking compared with 3\% for the OAK-D Pro W.
Most D435i failures were caused by scale divergence (18 runs), while the OAK-D Pro W had only three scale failures together with a small number of process and ATE failures. 
% The scale bias across the D435i IMU-based methods suggests that IMU--camera extrinsic calibration may have contributed to the performance gap, so the difference should not be attributed solely to sensor hardware.
A more detailed breakdown is provided in Appendix~\ref{app:failures}.

\begin{table}[t]
\centering
\caption{Failure summary for the two sensors across all benchmark runs.
``Process fail'' denotes an execution failure; ``No ATE'' indicates that a
trajectory was produced but could not be evaluated; ``Scale div.'' denotes
scale outside the accepted $0.7$--$1.5$ range; and ``Excursion'' denotes a
large transient trajectory error.}
\label{tab:failure_summary}
\resizebox{\columnwidth}{!}{%
\begin{tabular}{lrrrrrrr}
\toprule
Sensor & Runs & Usable & Process fail & No ATE & Scale div. & Excursion & Failed \\
\midrule
OAK-D Pro W     & 252 & 244 & 3 & 1 & 3  & 1 & 3\% \\
RealSense D435i & 252 & 228 & 0 & 0 & 18 & 6 & 10\% \\
\midrule
Total           & 504 & 472 & 3 & 1 & 21 & 7 & 6.3\% \\
\bottomrule
\end{tabular}%
}
\end{table}

We also report the computational costs of each algorithm to allow for algorithm selection during onboard deployment on embedded systems.
An Real-Time Factor (RTF) below 1
indicates that a method processes the sequence faster than real time on our benchmark workstation, while RTF above 1 indicates that processing is slower than the incoming data rate.
%These values should not be interpreted as direct predictions of performance on embedded hardware.
Although runtime scaling is hardware-dependent, the ordering of the algorithms by computational cost will translate to embedded systems.
%speedup between two methods on the benchmark workstation may not be preserved on platforms such as the Raspberry Pi 5.
%Nevertheless, reporting RTF together with CPU utilization and
%peak memory provides a common reference for computational demand and can help
%guide algorithm selection and onboard hardware design.

The results show a more consistent effect of the sensor configuration than of the algorithm choice. 
Across the meadow and above-canopy sequences, the OAK-D Pro W generally gives lower trajectory error and higher completion rates than the D435i, although the magnitude of the improvement varies by method. 
Previous sensor-level comparisons have also reported substantial differences between D435 and OAK-D Pro under different operating conditions~\cite{rustler2025empirical}. 
%However, this should not be interpreted as a controlled head-to-head comparison of the two cameras. 

There are several difference between the two sensors that contribute to a difference in accuracy. 
As shown in Table~\ref{tab:sensors}, the sensors differ not only in field of view and resolution, but also in stereo frame rate (30~Hz for the D435i versus 40~Hz for the OAK-D Pro W) and IMU sampling rate. 
These differences are important for VI-SLAM, since a higher update rate can provide greater temporal overlap between observations during rapid motion, thereby improving tracking performance. 
Furthermore, the wider field of view and higher stereo rate of the OAK-D Pro W will contribute to improved feature retention during motion.

\section{Data Collection Platform}

Our ForVis dataset was collected using a custom 5-inch quadrotor UAV designed for low-altitude forest flight and equipped with a Raspberry Pi 5 (RPi5) for onboard sensing and data logging. The platform builds on an improved ARK-RIG5 design from ARK Electronics and was selected with the consideration of USA's National Defense Authorization Act (NDAA) Section 889 compliance. Two camera systems were mounted on vibration-damping mounts to provide complementary visual and inertial measurements. Their specifications and configurations are summarized in Table~\ref{tab:sensors}. The mounts maintain a fixed relative transformation between the sensors and vehicle frame while reducing transmission of airframe vibration. Power for the onboard computer and cameras was supplied by the UAV power system through regulated converters; the OAK camera and RPi5 were supplied through a dedicated converter because of the camera's higher power requirement. Further details on the GNSS/RTK reference setup and its performance under forest canopy are provided in Appendix~\ref{app:gnss}.

% \begin{figure}[t]
%     \centering
%     \includegraphics[width=0.65\linewidth]{img/drone.png}
%     \caption{The UAV platform used for the ForVis dataset.}
%     \label{fig:uav_platform}
% \end{figure}

\begin{table}[t]
\centering
\caption{Sensor specifications for the ForVis dataset.}
\label{tab:sensors}
\resizebox{\linewidth}{!}{%
\begin{tabular}{lcc}
\hline
\textbf{Specification} & \textbf{RealSense D435i} & \textbf{OAK-D Pro Wide} \\
\hline
Stereo IR Resolution & $848\times480$ & $1280\times800$ \\
Stereo IR Rate       & 30 Hz          & 40 Hz \\
RGB Resolution       & $848\times480$ & $640\times400$ \\
RGB Rate             & 30 Hz          & 20 Hz \\
Gyroscope Rate       & 200 Hz         & 400 Hz \\
Accelerometer Rate   & 100 Hz         & 400 Hz \\
FOV (D/H/V)          & $95^\circ/87^\circ/58^\circ$ &
                        $150^\circ/127^\circ/79.5^\circ$ \\
Downward Tilt        & $\sim22^\circ$ ($12^\circ$ under canopy) &
                        $\sim33^\circ$ ($25^\circ$ under canopy) \\
\hline
\end{tabular}%
}
\end{table}

% \subsection{Dataset Analysis}

%The benchmark also characterizes the dataset. 
%Difficulty levels are more strongly associated with failure than with trajectory error.
% Across both sensors and all methods, failed runs increase from easy to medium to hard (6, 20, and 29 of 126), as do configurations with no usable run (2, 5, and 9 of 42). 
%Median error is not monotonic with difficulty, indicating that harder sequences primarily increase the risk of failure rather than reduce the accuracy of successful estimates. 
%Failure counts are therefore treated as a primary result.
%On the OAK, the best-to-worst method spread is 0.74–0.97 m under canopy, 0.98–1.38 m in the meadow, and 8.74–12.72 m above canopy. Above-canopy sequences therefore most clearly separate methods, whereas under-canopy sequences are least discriminative, with all methods within roughly 1 m. Thus, above-canopy sequences are best for method comparison, hard sequences for robustness evaluation, and under-canopy sequences for assessing the intended application.

\section{Conclusion} 
\label{sec:conclusion}

We presented ForVis, a real-world UAV dataset and benchmark for VI-SLAM in forest environments. 
The dataset combines synchronized stereo and inertial measurements from two sensors with flight-controller data across meadow, above-canopy, and under-canopy flights.
Benchmarking seven open-source systems shows that performance depends strongly on both the sensors and the flight conditions. 
Across all methods, the OAK-D Pro produces lower trajectory error and fewer failures than the D435i, because the sensors differ in field of view, resolution, frame rate, stereo baseline, and IMU sampling rate. 
%Difficulty is expressed primarily through estimator failure rather than steadily increasing error, 
The above-canopy sequences provides the strongest separation between algorithms, on the other hand under-canopy sequences challenge the algorithms the most.
Together, these results show that ForVis can advance the state-of-the-art VI-SLAM by challenging researchers to develop more accurate, robust or faster algorithms for real world deployment. 
%complementary evaluation of accuracy, robustness, and sensing configuration for visual-inertial navigation in realistic forest flight.

\section*{Acknowledgments}

This material is based upon work supported by the National Science Foundation
under Award No. 2416915 (E-RISE RII). The authors thank their collaborators
and lab members for their support in the development of the UAV platform.

%% Use plainnat to work nicely with natbib. 

\bibliographystyle{plainnat}
\bibliography{references}

\appendices
\section{Appendices}
\subsection{Additional Benchmark Results}
\label{app:benchmark_results}

We provide additional benchmark results in Table~\ref{tab:results}.

\begin{table*}[t]
  \centering
  \caption{Absolute trajectory error (SE(3), RMSE, metres) for seven methods on twelve
  sequences and two cameras. Each entry is the mean over three runs; runs whose scale
  diverged (outside $\times0.7$--$1.5$) or which threw a transient excursion are excluded,
  and a superscript gives the number of runs contributing where fewer than three survived.
  ``---'' means no run on that cell produced a usable trajectory. The six under-canopy
  columns are scored against a meter-grade GNSS reference and are not comparable with the
  six open-sky columns.}
  \label{tab:results}
  \resizebox{\textwidth}{!}{%
  \begin{tabular}{lcccccccccccc}
    \toprule
     & \multicolumn{3}{c}{Meadow} & \multicolumn{3}{c}{Above canopy} & \multicolumn{6}{c}{Under canopy} \\
    \cmidrule(lr){2-4}\cmidrule(lr){5-7}\cmidrule(lr){8-13}
    Method & 1\, & 2\, & 3\, & 1\, & 2\, & 3\, & 1\, & 2\, & 3\, & 4\, & 5\, & 6\, \\
    \midrule
    \multicolumn{13}{l}{\textit{OAK-D Pro W}} \\
    \texttt{orb\_slam3} & 0.58 & 0.66 & 0.27 & --- & 1.69$^{2}$ & 5.74 & 1.77 & 2.05 & 2.50 & 1.18 & 1.73 & 1.11 \\
    \texttt{okvis2\_x} & 0.43 & 0.25 & 0.28 & 0.76 & 1.24 & 6.29 & 1.66 & 2.31 & 2.48 & 1.36 & 1.93 & 1.70 \\
    \texttt{basalt} & 1.80 & 0.25 & 1.71 & 0.43 & 1.03 & 3.55 & 1.71 & 2.38 & 2.32 & 1.47 & 2.21 & 1.67 \\
    \texttt{stella\_vslam} & 0.64 & 1.45 & 1.10 & 1.94 & 2.54 & 14.31 & 1.74 & 2.23 & 2.50 & 1.33 & 2.01 & 1.76 \\
    \texttt{svo\_pro\_open} & 1.09 & 0.48 & 0.62 & 1.06 & 1.31 & 4.40 & 1.66 & 2.31 & 2.49 & 1.43 & 1.79 & 1.86$^{2}$ \\
    \texttt{openvins} & 0.38 & 0.59 & 0.30 & 1.48 & 1.54 & 8.78 & 1.79 & 1.97 & 2.20 & 1.27 & 2.25 & 1.46 \\
    \texttt{kimera\_vio} & 0.86 & 1.72 & 0.44 & 5.98 & 11.81 & --- & 2.59 & 2.05 & 3.30 & 1.49 & 2.94 & 1.74 \\
    \midrule
    \multicolumn{13}{l}{\textit{RealSense D435i}} \\
    \texttt{orb\_slam3} & 0.33 & 0.49 & 0.24 & 1.00 & 1.92 & 0.49 & 1.75 & 2.12 & 2.38 & 1.52 & 2.20 & 1.17 \\
    \texttt{okvis2\_x} & 0.42 & 0.37 & 0.46 & 0.78 & 1.25 & 6.20 & 1.78 & 2.73 & 2.41 & 1.74 & 2.17 & 1.97 \\
    \texttt{basalt} & 0.30 & 0.45 & 0.41 & 0.64 & 2.46 & 1.88 & 1.68 & 2.50 & 2.50 & 1.68 & 2.53 & 2.02 \\
    \texttt{stella\_vslam} & 1.87 & 2.92 & 2.85 & 8.86 & 8.46$^{1}$ & --- & 2.13 & 3.03 & 2.41 & 2.02 & 2.07 & 2.02 \\
    \texttt{svo\_pro\_open} & 0.63 & 0.64$^{2}$ & 0.75$^{2}$ & 0.48$^{1}$ & --- & 5.38 & 1.80 & 2.62 & 2.25 & 1.91 & 1.92$^{2}$ & 2.09 \\
    \texttt{openvins} & --- & 0.77 & 0.87 & 1.81 & 3.53 & --- & 2.14 & 2.05 & 2.20 & 2.49 & 4.99 & 1.62 \\
    \texttt{kimera\_vio} & 1.66 & 5.19 & 2.01 & 8.47$^{2}$ & 12.52$^{2}$ & --- & 2.17 & 2.17 & 2.78 & 1.31 & 2.11 & 1.94 \\
    \bottomrule
  \end{tabular}}
\end{table*}

\subsection{GNSS/RTK Reference Setup}
\label{app:gnss}

% Some of these details can be in the appendix
GNSS reference positioning was recorded using an ARK X20 RTK GPS equipped with a u-blox ZED-X20P all-band GNSS receiver. RTK correction data were obtained from a reference station located approximately 18~km from the flight site and delivered through the RTK2go NTRIP caster using an Internet connection provided by a smartphone hotspot. This correction link was used during all data-collection flights. Only RTK-fixed solutions (MAVLink \texttt{fix\_type = 6}) are accepted as GNSS ground truth~\cite{mavlink_gps_fix_type}. According to the ZED-X20P specifications, RTK positioning accuracy under suitable conditions is approximately $0.006~\mathrm{m}+1~\mathrm{ppm}$ horizontally and $0.01~\mathrm{m}+1~\mathrm{ppm}$ vertically~\cite{ublox_zedx20p}. Previous work has likewise shown that corrected low-cost multi-frequency GNSS receivers can achieve centimeter-level positioning under favorable conditions~\cite{robustelli2023low}.

The meadow and above-canopy flights maintained reliable RTK-fixed positioning for use as trajectory reference. Under the forest canopy, however, RTK-fixed solutions were intermittent and could not provide a continuous reliable reference. The source of these interruptions could not be isolated: canopy obstruction and multipath can degrade satellite measurements, while cellular connectivity and correction-stream availability can also affect RTK operation. RTK correction service availability was therefore monitored during data collection, but intermittent RTK fixes under canopy were not treated as ground truth. This is consistent with previous reports of degraded GNSS performance beneath tree canopy. Eren et al. reported horizontal RTK accuracies of approximately $1$--$12~\mathrm{cm}$ in repeated measurements under canopy~\cite{eren2025evaluation}. Consequently, RTK-fixed GNSS is used as the ground-truth trajectory for the open-sky sequences, whereas under-canopy GNSS is retained only as auxiliary positioning information and trajectory consistency is assessed using loop-closure constraints.

\subsection{Dataset Organization}

Each ForVis sequence contains synchronized stereo and inertial measurements
from the RealSense D435i and OAK-D Pro Wide, together with acquisition
metadata and, when available, PX4 flight-controller logs. Each camera
recording is provided as a standalone EuRoC-format dataset to simplify
use with existing VIO pipelines.

Both camera systems use the Raspberry Pi host clock and store timestamps in
Unix time with nanosecond resolution. The streams are therefore timestamp
aligned rather than hardware synchronized, and corresponding frames from the
two independent cameras may have a small temporal offset.

The complete directory structure, including calibration, metadata, optional
RGB/depth data, and flight-controller logs, is shown in
Fig.~\ref{fig:data_structure} in Appendix~\ref{app:data_structure}.
A conversion script will also be provided for generating ROS bags from the
released sequences.

\begin{table*}[t]
\centering
\caption{Failure breakdown by method across both sensors.}
\label{tab:failure_method}
\begin{tabular}{lrrrrrrr}
\toprule
Method & Runs & Usable & Process & No ATE & Scale & Excursion & Failed \\
\midrule
ORB-SLAM3    & 72 & 68 & 3 & 1 & 0  & 0 & 6\% \\
OKVIS2-X     & 72 & 72 & 0 & 0 & 0 & 0 & 0\% \\
Basalt       & 72 & 72 & 0 & 0 & 0 & 0 & 0\% \\
Stella-VSLAM & 72 & 67 & 0 & 0 & 5  & 0 & 7\% \\
SVO Pro      & 72 & 63 & 0 & 0 & 2  & 7 & 12\% \\
OpenVINS     & 72 & 66 & 0 & 0 & 6 & 0 & 8\% \\
Kimera-VIO   & 72 & 64 & 0 & 0 & 8 & 0 & 11\% \\
\bottomrule
\end{tabular}
\end{table*}

\begin{table*}[t]
\centering
\caption{Failure breakdown by environment across all methods and sensors.}
\label{tab:failure_environment}
\begin{tabular}{lrrrrrrr}
\toprule
Environment & Runs & Usable & Process & No ATE & Scale & Excursion & Failed \\
\midrule
Open meadow  & 126 & 121 & 0 & 0 & 3  & 2 & 4\% \\
Above canopy & 126 & 101 & 3 & 1 & 18 & 3 & 20\% \\
Under canopy & 252 & 250 & 0 & 0 & 0  & 2 & 1\% \\
\bottomrule
\end{tabular}
\end{table*}

\begin{table*}[t]
\centering
\caption{Failure breakdown by sequence across all methods and sensors.}
\label{tab:failure_sequence}
\begin{tabular}{llrrrrr}
\toprule
Sequence & Environment & Runs & Usable & Crash / No ATE & Guard & Failed \\
\midrule
mdw\_1 & Open meadow  & 42 & 39 & 0 & 3  & 7\% \\
mdw\_2 & Open meadow  & 42 & 41 & 0 & 1  & 2\% \\
mdw\_3 & Open meadow  & 42 & 41 & 0 & 1  & 2\% \\
abv\_1 & Above canopy & 42 & 36 & 3 & 3  & 14\% \\
abv\_2 & Above canopy & 42 & 35 & 1 & 6  & 17\% \\
abv\_3 & Above canopy & 42 & 30 & 0 & 12 & 29\% \\
can\_1 & Under canopy & 42 & 42 & 0 & 0  & 0\% \\
can\_2 & Under canopy & 42 & 42 & 0 & 0  & 0\% \\
can\_3 & Under canopy & 42 & 42 & 0 & 0  & 0\% \\
can\_4 & Under canopy & 42 & 42 & 0 & 0  & 0\% \\
can\_5 & Under canopy & 42 & 41 & 0 & 1  & 2\% \\
can\_6 & Under canopy & 42 & 41 & 0 & 1  & 2\% \\
\bottomrule
\end{tabular}
\end{table*}

\subsection{FAST Features}

% What is the message you want to give from FAST features?
% I do no think this paragraph is important
%%%%%%%%% FAST feature detections %%%%%%%%%%%%%%%%%%%%%%%%%%%%%%%%
The visual content also differs substantially across environments.
Under-canopy sequences contain the highest number of FAST features, while
above-canopy flights contain the fewest. For the OAK-D camera, the median
number of detected corners is 10,212--11,096 under canopy, compared with
5,805--7,104 in the meadow and 2,934--5,414 above canopy, reflecting the
closer and more textured structure below the canopy.

% % Not important? --> actually it is very important because it directly changes the accuracy.
% Delivered stereo frame rates reach 87.3--94.4\% of the requested rate, with
% most losses occurring as isolated frames; one 634~ms dropout is retained in
% the dataset.
% 
% 
% %% Details
% Propeller vibration causes severe accelerometer noise on the 5-inch
% airframe: using nominal manufacturer noise parameters, all tested methods
% diverge on every sequence. We therefore use increased accelerometer noise
% density and bias random-walk values, fixed across all methods for a given
% sensor. The same parameters are released with the benchmark harness.

\subsection{Failure Analysis}
\label{app:failures}

Table~\ref{tab:failure_method} shows clear differences in robustness across methods. SVO Pro has the highest failure rate at 12\%, followed by Kimera-VIO at 11\%, while OKVIS2-X and Basalt have the lowest at 0\%. Most failures are caused by scale divergence, particularly for Stella-VSLAM, SVO Pro, OpenVINS, and Kimera-VIO. ORB-SLAM3 is the only method with process and no-ATE failures but no scale divergence and excursion, while SVO Pro shows seven transient excursions. Overall, the table shows that failures are mostly due to trajectory quality rather than software crashes.

Table~\ref{tab:failure_environment} shows clear differences in robustness across environments. Above-canopy sequences have the highest failure rate at 20\%, compared with 4\% in open meadow and only 1\% under canopy. Most above-canopy failures are caused by scale divergence, with additional process, no-ATE, and excursion failures. Open-meadow failures are limited to scale divergence and excursions, while under-canopy sequences are highly robust, with only two excursion failures. Overall, the results indicate that the above-canopy environment is the most challenging condition for the evaluated methods.

Table~\ref{tab:failure_sequence} shows that failure rates also vary noticeably between individual sequences. The highest failure rate occurs on abv\_3 at 29\%, followed by abv\_2 at 17\% and abv\_1 at 14\%, confirming that the above-canopy sequences are the most challenging. In contrast, can\_1 to can\_4 have no failures, while can\_5 and can\_6 each have only a 2\% failure rate. Among the open-meadow sequences, mdw\_1 is the most difficult at 7\%, whereas mdw\_2 and mdw\_3 each have a 2\% failure rate. Overall, the sequence-level results are consistent with the environment-level trends.

\section{Dataset Structure}
\label{app:data_structure}

The dataset is organized into a EuRoC format with a script to convert EuRoC to rosbag if needed.
% \documentclass[10pt,a4paper]{article}
% \usepackage[margin=1in]{geometry}
% \usepackage{dirtree}
% \makeatletter
% \renewcommand{\DTcomment}[1]{%
%   \hspace*{\fill}%
%   \makebox[0.43\textwidth][l]{\DTstylecomment #1}%
% }
% \makeatother
% \usepackage{caption}

% \begin{document}

\begin{figure*}[t]
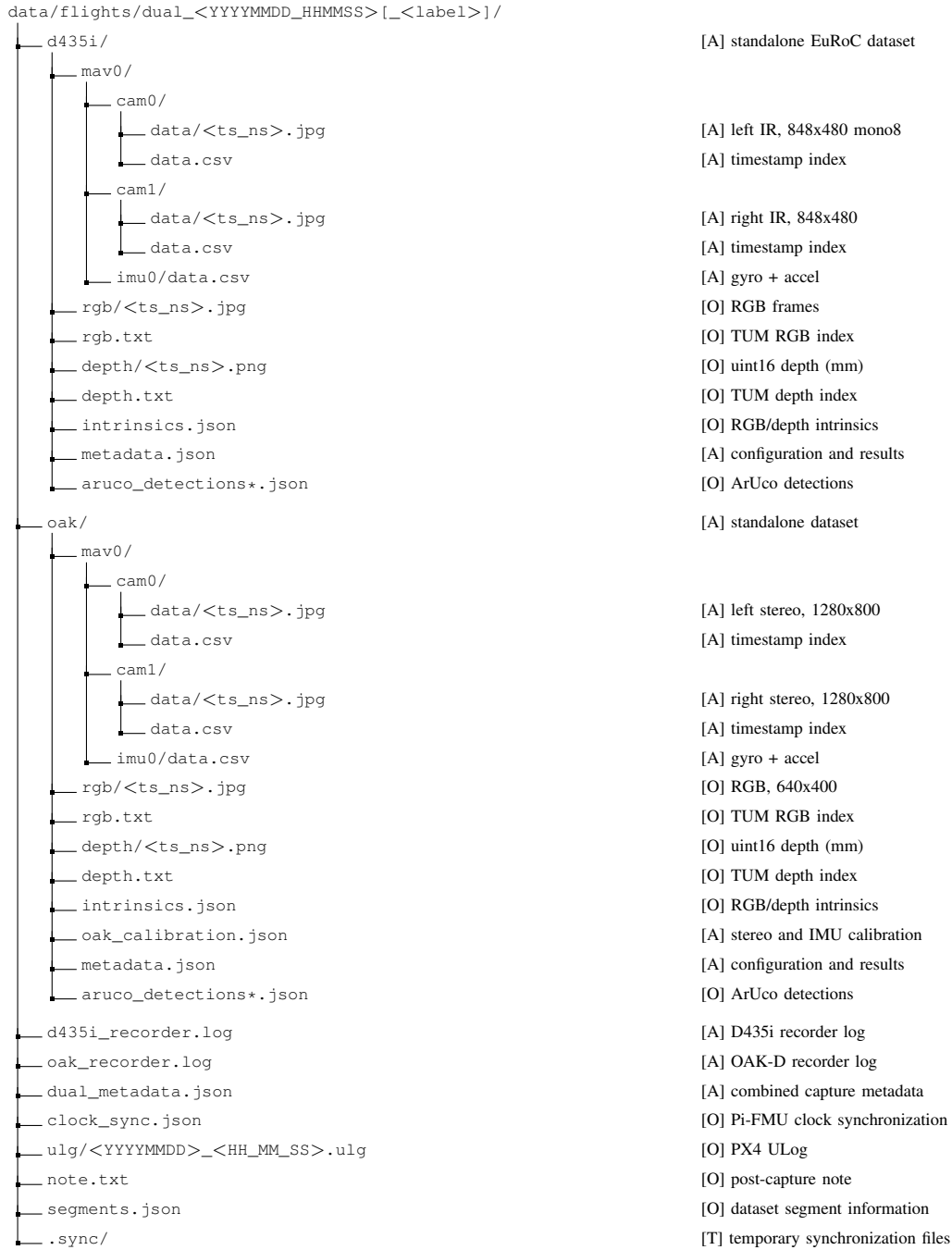

% \centering
\scriptsize
% Fix wide spacing and keep node text close to tree lines
% \setlength{\DTXoffset}{0.2em}
\dirtree{%
.1 data/flights/dual\_$<$YYYYMMDD\_HHMMSS$>$[\_$<$label$>$]/.
.2 d435i/\DTcomment{[A] standalone EuRoC dataset}.
.3 mav0/.
.4 cam0/.
.5 data/$<$ts\_ns$>$.jpg\DTcomment{[A] left IR, 848x480 mono8}.
.5 data.csv\DTcomment{[A] timestamp index}.
.4 cam1/.
.5 data/$<$ts\_ns$>$.jpg\DTcomment{[A] right IR, 848x480}.
.5 data.csv\DTcomment{[A] timestamp index}.
.4 imu0/data.csv\DTcomment{[A] gyro + accel}.
.3 rgb/$<$ts\_ns$>$.jpg\DTcomment{[O] RGB frames}.
.3 rgb.txt\DTcomment{[O] TUM RGB index}.
.3 depth/$<$ts\_ns$>$.png\DTcomment{[O] uint16 depth (mm)}.
.3 depth.txt\DTcomment{[O] TUM depth index}.
.3 intrinsics.json\DTcomment{[O] RGB/depth intrinsics}.
.3 metadata.json\DTcomment{[A] configuration and results}.
.3 \vspace{0.5em}aruco\_detections*.json\DTcomment{[O] ArUco detections}.
.2 oak/\DTcomment{[A] standalone dataset}.
.3 mav0/.
.4 cam0/.
.5 data/$<$ts\_ns$>$.jpg\DTcomment{[A] left stereo, 1280x800}.
.5 data.csv\DTcomment{[A] timestamp index}.
.4 cam1/.
.5 data/$<$ts\_ns$>$.jpg\DTcomment{[A] right stereo, 1280x800}.
.5 data.csv\DTcomment{[A] timestamp index}.
.4 imu0/data.csv\DTcomment{[A] gyro + accel}.
.3 rgb/$<$ts\_ns$>$.jpg\DTcomment{[O] RGB, 640x400}.
.3 rgb.txt\DTcomment{[O] TUM RGB index}.
.3 depth/$<$ts\_ns$>$.png\DTcomment{[O] uint16 depth (mm)}.
.3 depth.txt\DTcomment{[O] TUM depth index}.
.3 intrinsics.json\DTcomment{[O] RGB/depth intrinsics}.
.3 oak\_calibration.json\DTcomment{[A] stereo and IMU calibration}.
.3 metadata.json\DTcomment{[A] configuration and results}.
.3 \vspace{0.5em}aruco\_detections*.json\DTcomment{[O] ArUco detections}.
.2 d435i\_recorder.log\DTcomment{[A] D435i recorder log}.
.2 oak\_recorder.log\DTcomment{[A] OAK-D recorder log}.
.2 dual\_metadata.json\DTcomment{[A] combined capture metadata}.
.2 clock\_sync.json\DTcomment{[O] Pi-FMU clock synchronization}.
.2 ulg/$<$YYYYMMDD$>$\_$<$HH\_MM\_SS$>$.ulg\DTcomment{[O] PX4 ULog}.
.2 note.txt\DTcomment{[O] post-capture note}.
.2 segments.json\DTcomment{[O] dataset segment information}.
.2 .sync/\DTcomment{[T] temporary synchronization files}.
}
\caption{Directory structure of a ForVis flight sequence. [A] always generated, [O] optional, and [T] temporary.}
\label{fig:data_structure}
\end{figure*}

% \end{document}

\end{document}